%% file: RSPNet_Arxiv.tex
\documentclass[final]{cvpr}

\usepackage{times}
\usepackage{epsfig}
\usepackage{graphicx}
\usepackage{amsmath}
\usepackage{amssymb}

\input{def.tex}

\usepackage{algorithm}
\usepackage{algorithmic}
\usepackage{multirow}

\usepackage[pagebackref=true,breaklinks=true,colorlinks,bookmarks=false]{hyperref}

\begin{document}

\title{RSPNet: Relative Speed Perception for \\Unsupervised Video Representation Learning}

\author{
	Peihao Chen$^{1}$\thanks{This work was done when Peihao Chen was a research intern at Baidu.}~~\thanks{Equal contribution.}~~, Deng Huang$^{1\dagger}$~~, Dongliang He$^{2}$, Xiang Long$^{2}$, Runhao Zeng$^{1}$, \\
	Shilei Wen$^{2}$, Mingkui Tan$^{1}$\thanks{Corresponding author.} , Chuang Gan$^{3}$  \\
	$^1$South China University of Technology, $^2$Baidu Inc., $^3$MIT-IBM Watson AI Lab \\
	{\tt\small \{phchencs, im.huangdeng, runhaozeng.cs, ganchuang1990\}@gmail.com,} \\
	{\tt\small \{hedongliang01, longxiang, wenshilei\}@baidu.com, 
	mingkuitan@scut.edu.cn}
}
\maketitle

\begin{abstract}
	We study unsupervised video representation learning that seeks to learn both motion and appearance features from \textbf{unlabeled video} only, which can be reused for downstream tasks such as action recognition.
	This task, however, is extremely challenging due to: 1) the highly complex spatial-temporal information in videos; and 2) the lack of labeled data for training. Unlike the representation learning for static images, it is difficult to construct a suitable self-supervised task to well model both motion and appearance features. 
	More recently, several attempts have been made to learn video representation through \textit{video playback speed prediction}. However, it is non-trivial to obtain precise speed labels for the videos. More critically, the learnt models may tend to focus on motion pattern and thus may not learn appearance features well.
	In this paper, we observe that the relative playback speed is more consistent with motion pattern, and thus provide more effective and stable supervision for representation learning.
	Therefore, we propose a new way to perceive the playback speed and exploit the \textbf{relative speed} between two video clips as labels. In this way, we are able to well perceive speed and learn better motion features.
	Moreover, to ensure the learning of appearance features, we further propose an \textbf{appearance-focused} task, where we enforce the model to perceive the appearance difference between two video clips.
	We show that optimizing the two tasks jointly consistently improves the performance on two downstream tasks. 
	Remarkably, for action recognition on UCF101 dataset, we achieve 93.7\% accuracy without the use of labeled data for pre-training. Code and pre-trained models can be found at https://github.com/PeihaoChen/RSPNet.
\end{abstract}


\begin{figure}[!t]
	\centering
	\includegraphics[width=\linewidth]{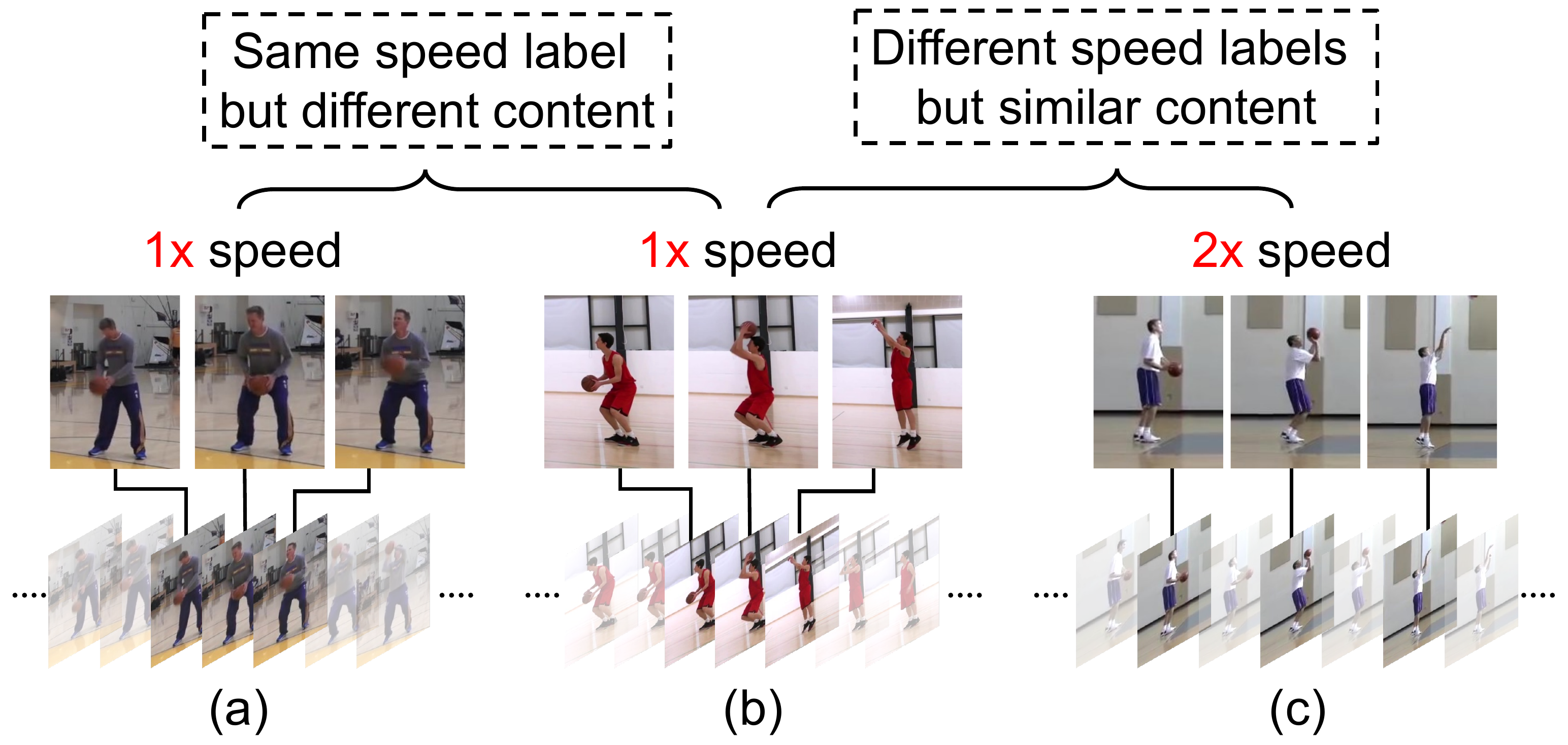}
	\caption{
		\textbf{An illustrative example of content-label inconsistency}.
		In existing speed perception-based methods~\protect\cite{speednet}, 
		1) both video clips (a) and (b) are labeled as 1x speed, \ie sampled consecutively, but the content of these two clips are dissimilar. The left player shoots more slowly and the middle player has finished shooting in the same time period. 
		2) Although clip (c) is labeled as 2x speed, \ie the sampling interval is set to 2 frames, it looks similar to the middle clip with different speed labels.
	}
	\label{fig:speed_label}
		\vspace{-4mm}
\end{figure}

	\section{Introduction}
\label{sec:intro}
	Video analysis~\cite{Huang2020Location} has been a prominent research topic in computer vision due to its vast potential applications, such as action recognition~\cite{LongGM0LW18,choi2019why}, event detection~\cite{GanWYYH15}, action localization~\cite{Chen2019,Zeng2019,Zeng2020Dense}, audio-visual scene analysis~\cite{Gan2019Self,Chen2020Generating,Gan2020Foley}, \textit{etc}. Compared with static images, 
	videos often contain more complex spatial-temporal contents and have a larger data volume, making it very challenging to annotate and analyze. How to learn effective video representations with a few annotations even without annotations is an important yet challenging task~\cite{GanSDG16,GanGLSG18,FanHGEGH18}. 

	Recently, unsupervised video representation learning, which seeks to learn appearance and motion features from unlabeled videos, has attracted great attention~\cite{playbackspeed,speednet,oops,GanYYYM16}.
	This task, however, is very difficult due to several challenges: 
	1) The downstream video understanding tasks, such as action recognition, rely on both appearance features (\eg texture and shape of objects, background scene) and motion features (\eg the movement of objects). It is difficult to learn representation for both appearance and motion simultaneously because of the complex spatial-temporal information in videos.
	2) It is difficult to mine effective supervision from unlabeled video data for representation learning.

Existing methods attempt to solve these challenges by designing pretext tasks to obtain pseudo labels for video representation learning. The pretext tasks include context prediction~\cite{HanXZ19}, playback rate perception~\cite{speednet}, temporal clip orders prediction~\cite{xu2019self}, \textit{etc}.
Among them, training models using playback speeds perception task achieves great success because models have to focus on the moving objects to perceive the playback speed~\cite{pace}. This helps models to learn representative motion features. Specifically, Benaim~\etal\cite{speednet} train a model to determine whether videos are sped up or not. Some works~\cite{oops,PRP,pace} try to predict the specific playback speed for each video.

However, these works suffer from two limitations. 
\textbf{First}, the playback speed labels used for pretext task can be imprecise because it may be inconsistent with motion content in videos. As shown in Figure \ref{fig:speed_label}, the clips with different labels (\ie different playback speeds) may look similar to each other.
The underlying reason is that different people often implement the same action at different speeds. Using such inconsistent speed labels for training may make it difficult to learn discriminative features. 
\textbf{Second}, perceiving speed mainly relies on motion content. It does not explicitly encourage models to explore appearance features which, however, are also important for video understanding.
Recently, instance discrimination task~\cite{wu2018unsupervised,he2019momentum} has shown its effectiveness for learning appearance features in image domain. However, how to extend it to video domain and combine it well with motion features learning is non-trivial.

To address the imprecise label issue in the above methods, we observe that the relative playback speed can provide more precise supervision for training. To this end, we propose a new pretext task that exploits relative playback speed as labels for perceiving speed, namely \textbf{Relative Speed Perception} (RSP). 
Specifically, we sample two clips from the same video and train a neural network to identify their relative playback speed instead of predicting the specific playback speed of each video clip. 
The relative playback speed label is obtained through the comparison between playback speeds of two clips from the same video (\eg 2x is faster than 1x). 
We observe that for the same video, the higher the playback speed, the faster the objects will move.
Consequently, such labels are independent of the original speed of objects in a video and can reveal the precise motion distinction between two clips. In this sense, the labels are more consistent with the motion content and can provide more effective supervision for representation learning.

Moreover, to encourage models to pay attention to learning appearance features, we follow the spirit of instance discrimination task in image domain and design an \textbf{Appearance-focused Video Instance Discrimination} (A-VID) task.
In this task, we require model to find out two clips sampled from the same video against a bunch of clips from other videos. 
Considering that different clips in the same video are often at the same speed, we propose a speed augmentation strategy, \ie randomizing the playback speed of each clip. Consequently, models cannot finish this task by simply learning speed information.
Instead, models tend to learn appearance features, such as background scene and the texture of objects, because these features are consistent along a video but vary among different videos.
We train models to finish RSP and A-VID tasks jointly using a two-branch architecture such that models are expected to learn both motion and appearance features simultaneously. We name our model as \textbf{RSPNet}.
Experimental results on three datasets show that the learnt features perform well on two downstream tasks, \ie action recognition and video retrieval.

To sum up, our contributions are as follows:

\begin{itemize}
	
	\item We propose a relative speed perception task for unsupervised video representation learning. In this sense, the labels are more consistent with the motion content and can provide more effective supervision for representation learning.
	
	\item We extend instance discrimination task to video domain and propose a speed augmentation strategy to make it focus more on exploring appearance content. In this way, we can combine it well with relative speed perception task to learn representation for both motion and appearance contents simultaneously.

	\item We verify the effectiveness of RSP and A-VID tasks for learning video representation on two downstream tasks and three datasets. Remarkably, without the need of annotation for pre-training, the action recognition accuracy on UCF-101 significantly outperforms the models supervised pre-trained on ImageNet (93.7\% \textit{v.s} 86.6\%).
\end{itemize}

\begin{figure*}[t]
	\centering
	\includegraphics[width=\linewidth]{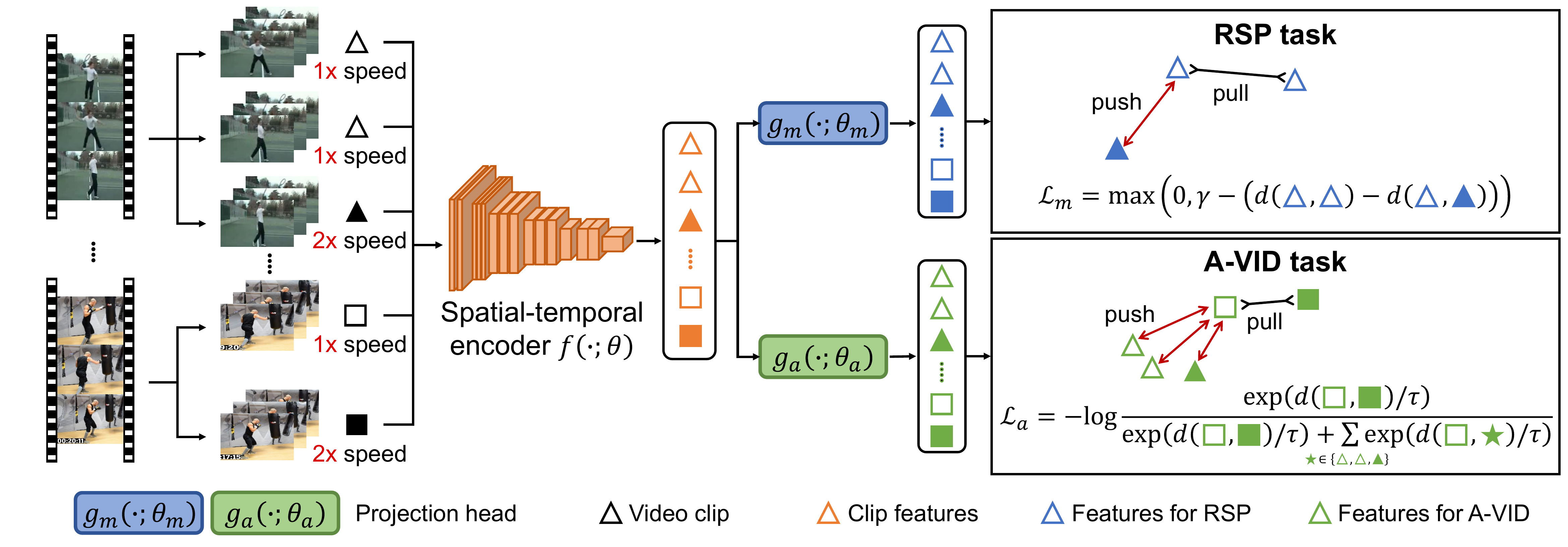}
	\caption{Illustration of the proposed self-supervised video representation learning scheme. Given a set of video clips with different playback speeds, we use a spatial-temporal encoder $f(\cdot;\theta)$ followed by two projection heads (\ie $g_m$ and $g_a$) to extract clip features for two pretext tasks. In the relative speed perception (RSP) task, we identify the relative playback speed between clips instead of predicting their specific playback speeds. In the appearance-focused video instance discrimination (A-VID) task, we distinguish video clips relying on the appearance contents.
	We formulate two tasks as a metric learning problem and use triplet loss $\mL_m$ and InfoNCE loss $\mL_a$ for training.}
	\label{fig:framework}
	\vspace{-0mm}
\end{figure*}	

	\vspace{-5mm}
	\section{Related work}
\paragraph{Unsupervised video representation learning.}
	In recent years, unsupervised video representation learning, which uses video itself as supervision, has become a popular topic~\cite{jing2020self}. 
	The existing methods learn representation through various carefully designed pretext tasks. Xu~\etal\cite{xu2019self} proposed video clip order prediction task for leveraging the temporal order of image sequences. Luo~\etal\cite{vcp} proposed video cloze procedure task by prediction the spatio-temporal operation applied on the video clips. Instead of focusing on RGB domain, Ng~\etal\cite{ng2018actionflownet} proposed a multitask learning model trained by estimating optical flow to learn motion representation. Since the video contains multiple frames, predicting future frames in latent space~\cite{oord2018representation} is also a effective task to learn visual representation.

	More recently, many works have been proposed to learn features through discriminating playback speeds. Epstein~\etal\cite{oops} try to predict whether a clip is sped up or not. Some works~\cite{pace,PRP} attempt to predict the specific playback speed of one clip. However, these works suffer from the imprecise speed label issue. Cho~\etal\cite{playbackspeed} design a method to sort video clips according to their playback speeds. However, they do not explicitly encourage model to learn appearance features. Our method makes use of relative speed to resolve the imprecise label issue. Moreover, we extent instance discrimination task~\cite{wu2018unsupervised} to video domain to encourage appearance learning.

\vspace{-5mm}

\paragraph{Metric Learning.}
	Metric learning~\cite{xing2002metric} aims to automatically construct task-specific distance metrics that compare two samples from a specific aspect. Based on such metric, the similar pairs of samples are pulled together and the dissimilar pairs of samples are pushed apart.
	It has achieved great success in many areas, \eg face recognition~\cite{tripletloss}, music recommendation~\cite{McFee2012music}, and person re-identification~\cite{yang2018reid}. 
	Recently, many works have successfully adopted metric learning for self-supervised representation learning~\cite{wu2018unsupervised,he2019momentum,tian2019contrastive}. They usually generate positive pairs by creating multiple views of each data and generate negative pairs by randomly choosing images/patches/videos. In this work, we aim to learn video representation by comparing two video clips using metric learning. Unlike the existing works, we propose to identify their speed distinction and appearance distinction to learn motion and appearance features from unlabeled data.

	\section{Proposed method}
\label{sec:Method}

\paragraph{Problem Definition.}
Let $\mV = \{ v_i \}_{i=1}^{N} $ be a video set containing $N$ videos. We sample a clip $\bc_i$ from a video with $s_i$ playback speed.
Unsupervised video representation learning aims to learn a spatial-temporal encoder $f(\cdot;\theta)$ to map video clip $\bc_i$ to their corresponding features $\bx_i$ that best describe the content in $\bc_i$.

This task is very challenging because of the complex spatial-temporal information in videos and the lack of annotations. 
It is difficult to construct supervision information from unlabeled videos $\mV$ to train a model to learn representation for both appearance and motion contents.
Recently, some existing unsupervised learning methods attempt to learn video representation through playback speed perception. However, most of them suffer from imprecise speed label issue and do not explicitly encourage models to learn appearance features. Consequently, the learnt features may not be suitable for downstream video understanding tasks such as action recognition and video retrieval.

\subsection{General scheme of RSPNet}
\label{sec:general_scheme}

In this paper, we observe that relative playback speed can provide more effective labels for representation learning. Thus, we propose a relative speed perception task, \ie predicting whether two clips are with the same speed or not, to resolve imprecise label issues and learn motion features. Moreover, we extend the instance discrimination task to video domain and propose a speed augmentation strategy to explicitly make models pay attention on exploring appearance features.
Considering the success of metric learning on representation learning~\cite{contrastiveloss,He2018scalelr}, we formulate these two tasks as metric learning, in which we seek to maximize the similarity of two clip features in positive pairs while minimizing the one in negative pairs. 

Formally, for \textbf{relative speed perception} task, 
instead of directly predicting playback speed $s_i$ for clip $\bc_i$, we propose to compare the speeds of two clips $\bc_i$ and $\bc_j$ that are sampled from the same video. 
Since the actions in $\bc_i$ (or $\bc_j$) are often implemented by the same subject, the motions in these two clips are similar when $s_i=s_j$ and are dissimilar when $s_i \neq s_j$.
In this sense, the relative speed labels are obtained through comparing $s_i$ and $s_j$ (\ie clips $\bc_i$ and $\bc_j$ are labeled as a positive pair when $s_i=s_j$ and otherwise negative). Such labels are more consistent with motion content in videos and reveal the precise motion distinction.
For \textbf{appearance-focused video instance discrimination} task, we enforce the model to predict whether two clips $\bc_i$ and $\bc_l$ are sampled from the same video. The intuition is that clips sampled from the same video often share similar appearance contents, which can be used as an important clue for distinguishing videos.
We also randomize the playback speed, \ie $s_i$ can be equal or not equal to $s_l$. In this way, models are encouraged to pay more attention on learning appearance features instead of finishing this task by learning playback speed information.

We use two individual projection heads $g_{m}(\cdot;{\theta_m})$ and $g_{a}(\cdot;{\theta_a})$ to map spatial-temporal features $\bc_i$ to $\bm_i$ and $\ba_i$ for two tasks, respectively. We train models on these two tasks jointly. The objective function is formulated as follows,	
\begin{equation}
	\label{eq:total}
	\mL(\mathcal{V}; \theta, \theta_a, \theta_m)=\mL_m(\mathcal{V}; \theta, \theta_m) + \lambda \mL_a(\mathcal{V}; \theta, \theta_a),
\end{equation}
where $\mL_m$ and $\mL_a$ denote the loss functions of each task, respectively and $\lambda$ is a fixed hyper-parameter to control the relative importance of each term. 
During inference for downstream tasks, we forward a video clip through the spatial-temporal encoder $f(\cdot;\theta)$ and obtain $\bx_i$ as its spatial-temporal features.
The schematic of our approach is shown in Figure~\ref{fig:framework}. In the following, we will introduce more details about two pretext tasks in Section~\ref{sec:task}.

\vspace{3mm}
\subsection{RSP and A-VID tasks}
\label{sec:task}

\paragraph{Relative speed perception.} This task aims to maximize the similarity of two clips with same playback speed and minimize the similarity of two clips with different playback speeds. Given a video, we sample 3 clips $\bc_i$, $\bc_j$ and $\bc_k$ with playback speeds $s_i$, $s_j$ and $s_k$, respectively, where $s_i = s_j \neq s_k $. We feed each clip into the spatial-temporal encoder $f(\cdot;\theta)$ followed by a projection head $g_m(\cdot;\theta_m)$ to get their corresponding features $\bm_i$, $\bm_j$, $\bm_k$. Dot product function $d(\cdot, \cdot)$ is used to measure the similarity between two clips. As the clips with the same playback speed share similar motion features, we expect their features can be closer compared with the clips with different playback speeds. We achieve this object by using a triplet loss~\cite{tripletloss} as follows,
\begin{equation}
	\label{eq:ranking}
	\mL_m(\mathcal{V}; \theta, \theta_m) = { {\rm max}(0,  \gamma - (p^{+} - p^{-})) },
\end{equation}
where $p^{+} = d(\bm_i, \bm_j)$, $p^{-} = d(\bm_i, \bm_k)$ and $\gamma >0$ is a certain margin. We desire that the similarity of a positive pair is larger than a negative pair by a margin $\gamma$.

\begin{algorithm}[t] 
	\small
	\caption{Training method of RSPNet}
	\label{algo:training}
	\begin{algorithmic}[1]
		\REQUIRE
		video set $\mV = \{ v_i \}_{i=1}^{N}$, \# negative pair for A-VID $K$.
		
		\STATE Initialize parameters $\theta, \theta_a, \theta_m$ for $f(\cdot;\theta), g_a(\cdot;\theta_a), g_m(\cdot;\theta_m)$, respectively
		
		\WHILE{no converge}
		
		\STATE Randomly sample a video $v^+$ from $\mV$, extract clips $c_i$, $c_j$, $c_k$ from $v^+$ with speed $s_i$, $s_j$, $s_k$, where $s_i = s_j \neq s_k$.
		
		\STATE Sample $K$ clips $\{ c_n \}_{n=1}^K$ from video set $\mV \setminus \{ v^+ \}$.
		
		\STATE Extract features $ \bx_i, \bx_j, \bx_k,$ and $\{ \bx_n \}_{n=1}^K$ from video clips  $c_i$, $c_j$, $c_k$, $\{ c_n \}_{n=1}^K$ using encoder $f(\cdot;\theta)$.
		
		\STATE // \emph{RSP task}
		\STATE Obtain features $ \bm_i, \bm_j, \bm_k $ from $\bx_i$, $\bx_j$, $\bx_k$ using $g_m(\cdot; \theta_m)$ .
		
		\STATE Compute $\mL_m$ using Equation (\ref{eq:ranking}).
		
		\STATE // \emph{A-VID task}
		\STATE Obtain features $ \ba_i, \ba_j, \{ \ba_n \}_{n=1}^K$ from $\bx_i$, $\bx_j$, $\{ \bx_n \}_{n=1}^K$ using  $g_a(\cdot; \theta_a)$
		
		\STATE Compute $\mL_a$ and $\mL$ using Equations (\ref{eq:nce}) and (\ref{eq:total}), respectively.
		
		\STATE Update parameters $\theta, \theta_a, \theta_m$ via stochastic gradient descent.
		
		\ENDWHILE
		
	\end{algorithmic}
\end{algorithm}

\begin{table*}[t]
	\centering
	\caption{Comparison of different pre-training settings on UCF101 and HMDB51 datasets. All models are pre-trained on the Kinetics-100 dataset except for the w/o pre-training setting. SP denotes speed prediction for each individual clip. VID denotes video instance discrimination without speed augmentation strategy.}
	\label{tab:task}
	\begin{tabular}{cccccccc}
		\hline
		\multicolumn{1}{c}{\multirow{2}{*}{Pre-training settings}}         & \multicolumn{3}{c}{UCF101} & &  \multicolumn{3}{c}{HMDB51} \\
		\cline{2-4}
		\cline{6-8}
		& TSM-18  & ResNet-18 & C3D   & & TSM-18  & ResNet-18  & C3D   \\ \hline
		w/o pre-training               & 49.7   & 42.3     & 59.0 & & 17.5   & 19.0      & 24.9 \\
		w/ RSP only       & 54.5   & 49.7     & 67.2 & & 26.5   & 25.9      & 29.4 \\ 
		w/ A-VID only  & 60.8   & 57.2     & 68.1 & & 30.2   & 31.1      & 35.1 \\ 
		SP + A-VID                   & 59.8   & 57.8     & 70.9 & & 29.7   & 30.7      & 35.1 \\ 
		RSP + VID  & 57.5   & 54.2     & 70.8 & & 30.1   & 29.9      & 34.5 \\ \hline
		RSP + A-VID (Ours)      & \textbf{61.2} & \textbf{60.2} & \textbf{71.5} & & \textbf{32.2} & \textbf{32.6} & \textbf{36.3} \\
		\hline
	\end{tabular}
\end{table*}

\paragraph{Appearance-focused video instance discrimination.} 
To explicitly encourage models to learn appearance features, we propose an A-VID task to further regularize the learning process. Motivated by the fact that different clips from the same video are always with similar spatial information, we extent the contrastive learning in the image domain~\cite{wu2018unsupervised} to the video domain. Specifically, we sample two clips $\bc_i$ and $\bc_j$ from the same  randomly selected video $v^+$ and $K$ clips $\{\bc_n\}_{n=1}^{K}$ from $K$ videos in subset $\mV \setminus v^+$. After that, we feed each clip into the spatial-temporal encoder $f(\cdot;\theta)$ followed by a projection head $g_a(\cdot;\theta_a)$ and get their corresponding features. The encoder $f(\cdot;\theta)$ share weights with the encoder in RSP task while the weights of projection head $g_a(\cdot;\theta_a)$ is independent of $g_m(\cdot;\theta_m)$. We consider ($\bc_i$, $\bc_j$) as positive pair and ($\bc_i$, $\bc_n$) as negative pair. We further apply the InfoNCE loss~\cite{he2019momentum} as the training loss:
\begin{equation}
	\label{eq:nce}
	\mL_a(\mathcal{V}; \theta, \theta_a) = - {\rm log} \frac{ q^{+}} 
	{q^{+} + \sum_{n=1}^{K} q^{-}_{n}}, 
\end{equation}
where $q^{+} = {\rm exp}(d(\ba_i,\ba_j) / \tau)$, $q^{-}_{n} = {\rm exp}(d(\ba_i,\ba_n) / \tau)$, and $\tau$ is a temperature hyper-parameter~\cite{wu2018unsupervised} which affects the concentration level of distribution. Optimizing Equation (\ref{eq:nce}) will pull closer the positive pairs while push away the negative pairs.

An underlying question is that how to sample these video clips. A naive solution is to sample all clips at the same playback speed. In this sense, clips $\bc_i$ and $\bc_j$ will share similar motion features while the motion features in $\bc_i$ and $\bc_n$ are dissmilar. This may provide clues for models to find out whether any two clips are from the same video or not. To encourage models to pay more attention on learning appearance features, we propose a speed augmentation strategy. Concretely, we randomize the playback speed of each clip, \ie $s_i$, $s_j$, and $s_n$ are randomly selected from possible playback speeds, such that the motion features cannot provide effective clues for this task. 
In this way models have to focus on learning other informative features, including background and object appearance for discriminating video instance. 
The training method is shown in Algorithm~\ref{algo:training}.

	\section{Experiments}

\paragraph{Datasets.}
We pre-train models on the training set of Kinetics-400 dataset~\cite{carreira2017quo}, which consists of around 240K training videos with 400 human action classes. Each video lasts about 10 seconds. To reduce training costs in ablation studies, we build a lightweight dataset, namely Kinetics-100, by selecting 100 classes with the least disk size of videos from Kinetics-400.
UCF101~\cite{ucf101} dataset consists of 13,320 videos from 101 realistic action categories on YouTube. HMDB51~\cite{hmdb51} dataset consists of 6,849 clips from 51 cation classes.
Compared with UCF101 and HMDB51, Something-Something-V2 (\emph{Something-V2}) dataset~\cite{something} contains 220,847 videos with 174 classes and focuses more on modeling temporal relationships~\cite{LinGH19,ZhouAOT18}.

\paragraph{Pre-training details.}
We instantiate the projection head as a fully connected layer with 128 output dimension. After pre-training, we drop the projection heads and use the features before them for downstream tasks.
Unless otherwise stated, we sample 16 consecutive frames with 112 $\times$ 112 spatial size for each clip following Kim~\etal\cite{3dpuzzle}.
Clips are augmented by using random cropping with resizing, random color jitter and random Gaussian blur~\cite{chen2020simple}.
We use SGD as the optimizer with a mini-batch size of 64. We train the model for 200 epochs by default.
The learning rate policy is linear cosine decay starting from 0.1.
Following He~\etal\cite{he2019momentum}, we set $\tau = 0.07$, $K = 16384$, $\gamma = 0.15$ and $\lambda=1$ for Equations  (\ref{eq:total}), (\ref{eq:ranking}) and (\ref{eq:nce}). 
All videos are with 25 fps.
The possible playback speed $s$ for clips in this paper is set to 1x (\ie sampling frames consecutively) and 2x (\ie sampling interval is set 2 frames).

\paragraph{Fine-tuning details.}
We fine-tune our RSPNet on UCF101, HMDB51, and \emph{Something-V2} with labeled videos for action recognition. We train for 30, 70, 50 epochs on these datasets, respectively, with a learning rate of 0.01. Following Xu~\etal\cite{xu2019self}, we initialize the models with the weights from the pre-trained RSPNet except for the newly appended fully-connected layer with randomly initialized weights.

\begin{table*}[ht]
	\centering
	\caption{Comparison with other unsupervised methods on UCF101 and HMDB51 datasets. We show the backbone architecture and the pre-training dataset of each method. *We pre-train the model for 1000 epochs.}
	\begin{tabular}{ccccc}
		\hline
		Method                & Architecture & Pre-train Dataset      & UCF101 & HMDB51 \\ 
		\hline
		Shuffle\&Learn~\cite{misra2016shuffle}  &  CaffeNet   & UCF101       & 50.2   & 18.1   \\
		CMC~\cite{tian2019contrastive} & CaffeNet  & UCF101       & 59.1   & 26.7   \\
		OPN~\cite{lee2017unsupervised} &  VGG      & UCF101       & 59.8   & 23.8   \\
		VCP~\cite{vcp}               & C3D     & UCF101       & 68.5   & 32.5   \\
		PSP~\cite{playbackspeed}     & R(2+1)D     & UCF101       & 74.8   & 36.8   \\
		ClipOrder~\cite{xu2019self}  & R(2+1)D     & UCF101       & 72.4   & 30.9   \\
		PRP~\cite{PRP}               & R(2+1)D     & UCF101       & 72.1   & 35.0   \\
		3D ST-Puzzle~\cite{3dpuzzle} & C3D          & Kinetics-400 & 60.6   & 28.3   \\
		MAS~\cite{WangJBHLL19}       &  C3D        & Kinetics-400 & 61.2   & 33.4   \\
		3D ST-Puzzle~\cite{3dpuzzle} & ResNet-18  & Kinetics-400 & 65.8   & 33.7   \\
		3DRotNet~\cite{jing2018self} & ResNet-18    & Kinetics-400 & 66.0   & 37.1   \\
		DPC~\cite{HanXZ19}           & ResNet-18  & Kinetics-400 & 68.2   & 34.5   \\
		MemDPC~\cite{han2020memdpc}  & ResNet-34       & Kinetics-400 & 78.7   & 41.2   \\
		Pace~\cite{pace}             & R(2+1)D     & Kinetics-400 & 77.1   & 36.6   \\
		CBT~\cite{sun2019learning}   & S3D          & Kinetics-600 & 79.5   & 44.6   \\
		CoCLR~\cite{han2020coclr}    & S3D       & Kinetics-400 & 87.9   & 54.6   \\
		SpeedNet~\cite{speednet}     & S3D-G       & Kinetics-400 & 81.1   & 48.8   \\

		\hline
		\multirow{2}{*}{Fully supervised} & S3D-G & ImageNet & 86.6  &  57.7  \\
		& S3D-G & Kinetics-400    & 96.8  &  75.9 \\
		\hline
		\multirow{5}{*}{RSPNet (Ours)} & C3D       & Kinetics-400 &  76.7  & 44.6     \\
		& ResNet-18        & Kinetics-400 &  74.3  &   41.8     \\ 
		& R(2+1)D      & Kinetics-400 &  81.1  &   44.6     \\ 
		& S3D-G        & Kinetics-400 &      89.9  &   59.6      \\ 
		& S3D-G        & Kinetics-400 &  \textbf{93.7}*  & \textbf{64.7}*       \\ 
		\hline
	\end{tabular}
	\label{tab:sota}
\end{table*}

\subsection{Ablation studies}

\paragraph{Effectiveness of two pretext tasks.}
In this paper, we propose two tasks, namely RSP and A-VID, to learn video representation. To verify the effectiveness of each task, we pre-train models using either RSP or A-VID on three backbone networks.  

In Table \ref{tab:task}, compared with training from scratch, using RSP or A-VID task for pre-training significantly improves the action recognition performance on UCF101 and HMDB51 datasets, which demonstrates models learn useful clues for action recognition through pre-training on our designed pretext task.
The improvement brought by A-VID task is relatively larger than pair-wise speed discrimination. The underlying reason is that UCF101 and HMDB51 datasets focus more on modeling appearance information compared with temporal relationship~\cite{LinGH19}. The models pre-trained on A-VID are more sensitive to object appearance and background scene while models pre-trained on RSP are more sensitive to the movement of objects.
When we pre-trained models on both task jointly, we achieve the best results on all three models. Compared with w/o pre-training setting, we achieve relative improvement of 11.5\%, 17.9\%, and 12.5\% on UCF101 and 14.7\%, 13.6\%, and 11.4\% on HMDB51 in top-1 accuracy. This demonstrates that the two pretext tasks are complementary to each other and are effective for learning video representation.

\paragraph{Does relative speed perception help?}
As discussed in Section \ref{sec:intro}, we train models to perceive relative speed of two clips to resolve the imprecise speed label issue. Here, we implement a variant of our method by replacing RSP with directly predicting speed of each clip (\ie 1x or 2x speed). We formulated it as a classification problem and use a cross-entropy loss to optimize it following Wang~\etal\cite{pace}. We denote this task as speed prediction (SP). Table \ref{tab:task} shows that exploiting relative speed as labels consistently improve the performance on three backbone networks and on two datasets compared with directly using playback speed of each clip (SP + A-VID \textit{v.s} RSP + A-VID). These results demonstrate that relative speed labels are more consistent with the motion content and help models to learn more discriminative video features. 

\paragraph{Does speed augmentation help?}
Instead of naively extend instance discrimination task from image domain to video domain, we propose to randomize the speed of each clip. To verify its effectiveness, we implement a variant by dropping speed augmentation. We denote it as VID as it is not appearance-focused. Table \ref{tab:task} shows that the speed augmentation strategy significantly improve the performance (RSP + VID \textit{v.s} RSP + A-VID). The reason is that the speed augmentation strategy make the VID task become speed-agnostic. In this way, models are encouraged to pay more attention on learning appearance features. Together with the motion features learnt from RSP task, models can extract more discriminative representation for appearance and motion, which are both important for action recognition.

\subsection{Evaluation on action recognition task}

\paragraph{Performance on UCF101 and HMDB51.}
We compare our method with the state-of-the-art self-supervised learning methods in Table~\ref{tab:sota}. We report top-1 accuracy on UCF101 and HMDB51 datasets together with the backbone and pre-training dataset.
As the prior works use different backbone networks for experiments, we report results using the same settings as theirs for fair comparisons. 

Our RSPNet achieves the best results on all backbone networks over two datasets. 
Specifically, with C3D, our method outperforms MAS~\cite{WangJBHLL19} by a large margin (76.7\% \textit{v.s} 61.2\% on UCF101 and 44.6\% \textit{v.s} 33.4\% on HMDB51). 
With ResNet-18, our method outperforms DPC~\cite{HanXZ19} by 6.1\% and 7.3\% absolute improvement on two datasets, respectively.
With R(2+1)D, our RSPNet improves accuracy from 77.1\% to 81.1\% on UCF101 and from 36.6\% to 44.6\% on HMDB51.
For S3D-G, we follow SpeedNet~\cite{speednet} to use video frames with 224 $\times$ 224 as input for pre-training and fine-tuning. Under the same settings, our RSPNet increase the accuracy from 81.1\% to 89.9\% on UCF101 and from 48.8\% to 59.6\% on HMDB51. 

\begin{figure}[t]
	\centering
	\includegraphics[width=\linewidth]{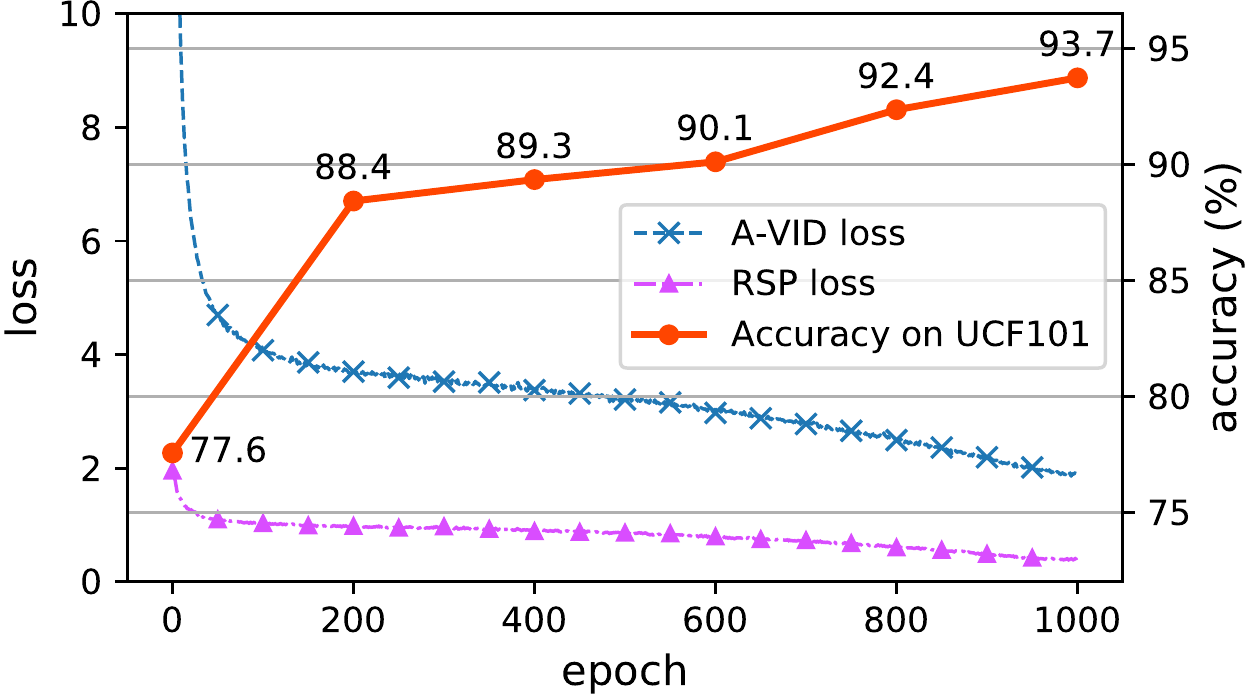}
	\caption{Pre-training losses of two pretext tasks and Top-1 accuracy of UCF101 after fine-tuning.
		We pre-train S3D-G model on K-400 for 1000 epochs and report the results every 200 epochs.
	}
	\label{fig:curve_of_epoch}
\end{figure}

When we train longer (\ie 1000 epochs), we can further improve the tkeop-1 accuracy to 93.7\% and 64.7\% for two datasets, respectively. In Figure \ref{fig:curve_of_epoch}, we show the curve of pre-training losses and the performance on UCF101 for S3D-G model using different checkpoints. As the losses decrease, the performance for downstream task increases consistently. This demonstrates the effectiveness of the proposed RSP and A-VID tasks.The model does learn semantic representation to solve them instead of learning trivial solution.
Remarkably, without the need of any annotation for pre-training, our RSPNet outperforms the ImageNet supervised pre-trained variant (93.7\% \textit{v.s} 86.6\%) and achieve close performance to the Kinetics supervised pre-trained model (96.8\%).

\paragraph{Performance on \emph{Something-V2}.}
We compare our RSPNet with supervised learning methods on \emph{Something-V2}, a challenging dataset in which temporal information is essential~\cite{LinGH19}.
Following the settings in Lin~\etal\cite{LinGH19}, we train models for 50 epochs and set the initial learning rate to 0.01 (decays by 0.1 at epoch 20 and 40). 
For the supervised pre-trained models, the ResNet-18 and S3D-G are pre-trained on K-400 dataset, and C3D is pre-trained on Sport-1M dataset~\cite{KarpathyTSLSF14}. 
Both K-400 and Sport-1M are large-scale datasets with manually annotated action labels, and thus the supervised pre-trained models are strong baselines for our unsupervised pre-trained RSPNet.

In Table~\ref{tab:something}, despite not using manual annotation, RSPNet consistently increases the accuracy compared with the random initialized models on three backbone architectures. Surprisingly, RSPNet even outperforms the supervised pre-trained model on ResNet-18 and C3D, increasing from 43.7\% to 44.0\% and from 47.0\% to 47.8\%, respectively. It shows the benefits of the discriminative features learnt from the proposed two pretext tasks.

\begin{table}[t]
	\centering
	\caption{Performance comparison on \emph{Something-V2}. }
	\label{tab:something}
	\begin{tabular}{ccccc}
		\hline
		& ResNet-18 & C3D & S3D-G \\ \hline
		w/o pre-training        &   42.1  & 45.8  & 51.2   \\ 
		Fully supervised          &   43.7  & 47.0  & \textbf{56.8} \\ \hline
		Unsupervised (Ours)    &\textbf{44.0}    & \textbf{47.8}  & 55.0 \\
		\hline
	\end{tabular}
\end{table}

\begin{table*}[ht]
	\centering
	\caption{Video retrieval results on UCF101, measured by top-$k$ retrieval accuracy (\%).}
	\label{tab:retri_acc}
	\begin{tabular}{ccccccc}
		\hline
		\multirow{2}[0]{*}{Method} & \multirow{2}[0]{*}{Architecture} &
		\multicolumn{5}{c}{Top-$k$} \\
		\cline{3-7}
		& & $k=1$    & $k=5$    & $k=10$   & $k=20$   & $k=50$     \\ 
		\hline
		OPN~\cite{lee2017unsupervised}      & OPN               & 19.9 & 28.7 & 34.0 & 40.6 & 51.6   \\
		Buchler~\etal\cite{buchler2018improving}  & CaffeNet          & 25.7 & 36.2 & 42.2 & 49.2 & 59.5   \\
		ClipOrder~\cite{xu2019self}    & R3D               & 14.1 & 30.3 & 40.0 & 51.1 & 66.5   \\
		SpeedNet~\cite{speednet} & S3D-G             & 13.0 & 28.1 & 37.5 & 49.5 & 65.0   \\
		VCP~\cite{vcp}       & R(2+1)D           & 19.9  & 33.7  & 42.0 & 50.5  & 64.4   \\
		Pace~\cite{pace}     & C3D               & 31.9  & 49.7  & 59.2 & 68.9  & 80.2   \\
		\hline
		\multirow{2}{*}{RSPNet (Ours)}    
		& C3D               & 36.0           & 56.7          & 66.5          & 76.3           & 87.7   \\
		& ResNet-18         & \textbf{41.1}  & \textbf{59.4} & \textbf{68.4} & \textbf{77.8}  & \textbf{88.7}   \\
		
		\hline
	\end{tabular}
\end{table*}

\subsection{Evaluation on video retrieval task}
Given a query video, we use the nearest neighbor search to retrieve relevant videos based on the cosine similarity of their features. 
Specifically, following previous works~\cite{speednet,xu2019self}, we evenly sample 10 clips for each video and take the output of the last convolutional layer in spatial-temporal encoder as clip-level features. Then, we perform spatial max-pooling on each clip and average-pooling over 10 clips to obtain a video-level feature vector. We use the video in testing set to retrieve the videos in training set.
We evaluate our method on the split 1 of UCF101 dataset and apply the top-$k$ accuracies ($k$=1, 5, 10, 20, 50) as evaluation metrics.
Our RSPNet is pre-trained on K-400 dataset.

From Table~\ref{tab:retri_acc}, our method outperforms state-of-the-arts by a large margin under different values of $k$. 
For example, our method achieves much better performance than Pace~\cite{pace} under all values of $k$ using the same C3D backbone. With ResNet-18 as backbone network, we can achieve better retrieval performance. These imply that the proposed pretext tasks help us to learn more discriminative features for video retrieval tasks.

\begin{figure}[t]
	\centering
	\includegraphics[width=\linewidth]{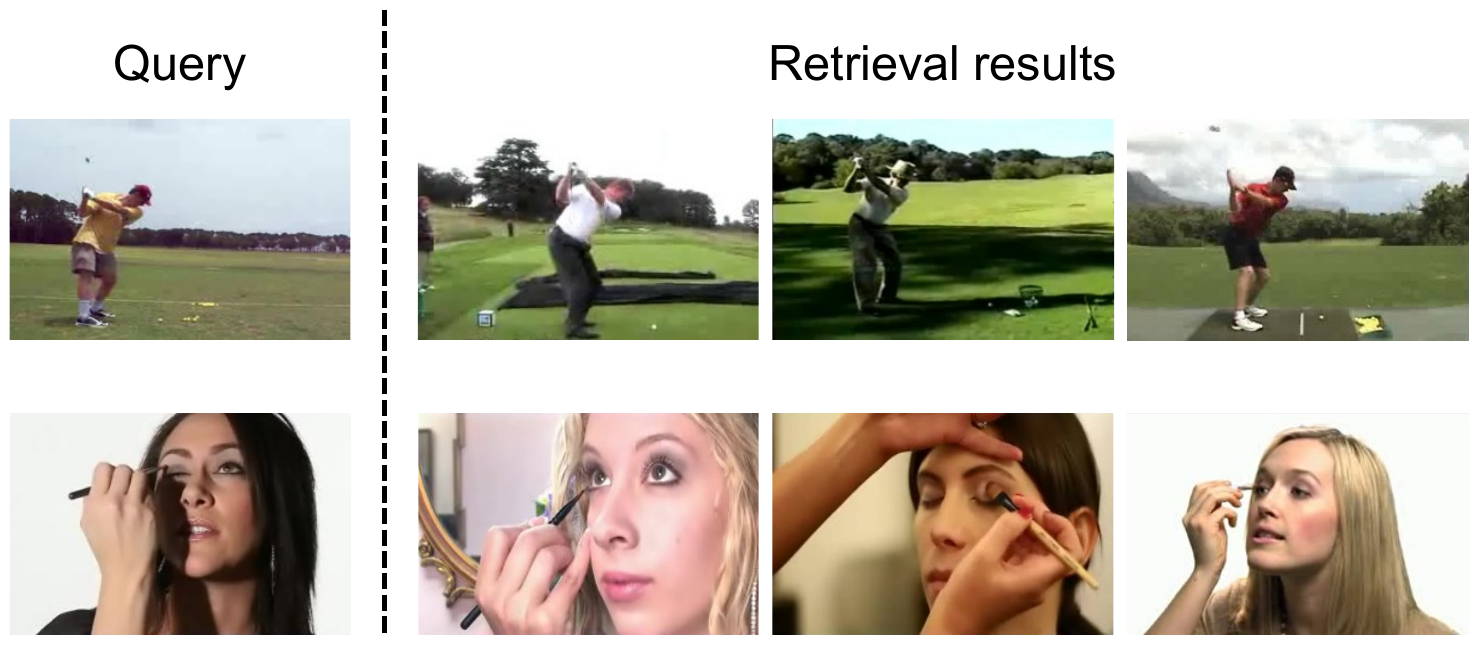}
	\caption{Qualitative examples of video retrieval.}
	\label{fig:retrieval_viz}
\end{figure}

We further provide some retrieval results in Figure~\ref{fig:retrieval_viz} as a qualitative study. For the two query clips, we successfully retrieve highly relevant videos with very similar appearance and motion. This implies that our method is able to learn both meaningful appearance and motion features for videos.

\subsection{RoI visualization} 
From Section \ref{sec:general_scheme}, we formulate two pretext tasks as metric learning, which seeks to maximize the similarity of the positive pair.  To better understand the clues learnt for the two pretext tasks, we visualize the region of interest (RoI) that contributes most to the similarity score using the class-activation map (CAM) technique \cite{zhou2016learning}.
We will describe the technical details of visualization followed by the results analysis.

In our RSPNet, we calculate the similarity $s$ between video clip features $\bx_i$ and $\bx_j$ using cosine distance, \ie $s = (\bW_j \bx_j)^\top (\bW_i \bx_i) = ((\bW_j \bx_j)^\top \bW_i) \bx_i$, where $\bW_i \in \mathbb{R}^{128\times C}$ and $\bW_j \in \mathbb{R}^{128\times C}$ are the parameters for the projection head $g_m$ (or $g_a$). The features $\bx_i$ is average pooled from the last convolutional features $\bF_i \in \mathbb{R}^{C\times H\times W\times T}$ of the spatial-temporal encoder, where $C$ is the number of channels and $H, W, T$ are spatial-temporal sizes. 
In analogy with CAM~\cite{zhou2016learning}, the similarity activation maps $\bM_s \in \mathbb{R}^{H\times W \times T}$ of clip $\bc_i$ for similarity score $s$ can be defined as
\begin{equation}\label{eq:ourVisualize}
\bM_s = ((\bW_j \bx_j)^\top \bW_i)  \bF_i.
\end{equation}
Such similarity activation maps indicate the salient regions of clip $\bc_i$ that are used by models to figure out whether the two clips are positive pair. We can also obtain activation maps of clip $\bc_j$ in a similar manner. More details can refer to Zhou~\etal\cite{zhou2016learning}.

Although both RSP and A-VID pretext tasks are based on the same features $\bF_i$, we use two independent projection heads $g_m$ and $g_a$ to map $\bF_i$ to different 128-D embedding spaces, as shown in Figure~\ref{fig:framework}. Thus, the parameters of linear layers, \ie $\bW_i$ and $\bW_j$, for two pretext tasks are different. Consequently, the activation maps can be different and models can focus on learning different clues for completing each specific pretext task.

\begin{figure}[]
	\centering
	\includegraphics[width=\linewidth]{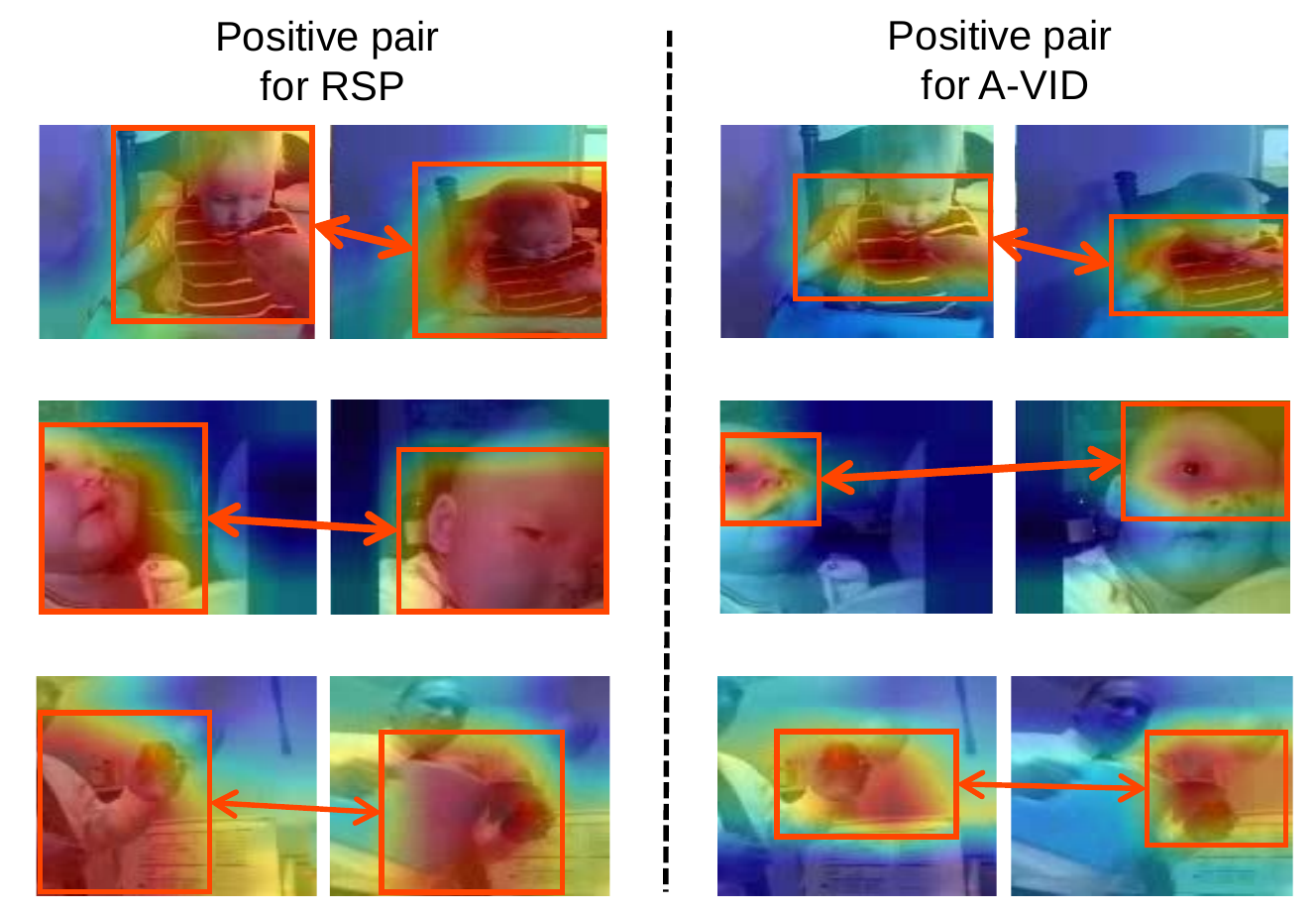}
	\caption{Visualization of RoI learnt for RSP and A-VID.
		Our model focuses on the regions containing rich motion and appearance information for two pretext tasks, respectively. 
		We outline the area where the heatmap is higher than a threshold with a rectangle.
	}
	\label{fig:vis_app_motion}
\end{figure}

In Figure \ref{fig:vis_app_motion}, we show the heatmaps in three positive clip pairs. We use the middle frame to represent a clip to visualize the heatmap. 
For the RSP task, the heatmaps tend to cover the whole region of actions, which provides rich information for perceiving the relative speed.
For the A-VID task, models tend to focus on small but discriminative regions (\eg the striped clothes and the eyes of a  baby in two pair samples, respectively) to identify two clips in the same video. One interesting finding is that the models are able to adaptively localize the same object even though they appear in different locations of a frame. This may provide a new perspective for person re-identification and we leave it for futher work.

\section{Conclusion}
In this paper, we have proposed an unsupervised video representation learning framework named RSPNet. 
We train models to perceive relative playback speed for learning motion features by using relative speed labels to resolve the imprecise speed label issue. Also, we extend instance discrimination task to video domain and propose a speed augmentation strategy to make models focus on learning appearance features.
Extensive experiments show that the features learnt by RSPNet perform better on action recognition and video retrieval downstream tasks. 
Visualization of RoI implies that RSPNet can focus on discriminative area for two tasks.

{\small
	\bibliographystyle{ieee_fullname}
	\bibliography{SSL}
}

\end{document}

%% file: def.tex
\def\etal{{\em et al.\/}\, }

\def\ie{\mbox{\textit{i.e.}, }}
\def\eg{\mbox{\textit{e.g.}, }}

\def\mL{{\mathcal L}}

\def\mV{{\mathcal V}}

\DeclareMathAlphabet\mathbfcal{OMS}{cmsy}{b}{n}

\def\0{{\bf 0}}
\usepackage{ dsfont }
\def\1{\mathds{1}}

\def\bF{{\bf F}}

\def\bM{{\bf M}}

\def\bW{{\bf W}}

\def\ba{{\bf a}}

\def\bc{{\bf c}}

\def\bm{{\bf m}}

\def\bx{{\bf x}}

\def\bx{{\bf x}}

\def\bW{{\bf W}}

\def\bc{{\bf c}}

\usepackage{ntheorem}

\newtheorem*{*thm}{Theorem}

\newtheorem*{*lemma}{Lemma}

\usepackage{enumitem}